\documentclass[letterpaper, 10 pt, conference]{ieeeconf}  

\IEEEoverridecommandlockouts                              

\usepackage{amsthm}
\usepackage{graphicx}
\usepackage{caption}
\usepackage{subcaption}
\usepackage[linesnumbered,ruled,vlined]{algorithm2e}
\usepackage{algorithmicx}
\usepackage{algpseudocode}
\usepackage{amsmath}
\theoremstyle{plain}
\usepackage{mathtools}
\usepackage{epsfig}
\usepackage{amssymb}
\usepackage{epstopdf}
\usepackage{multicol}
\usepackage{multirow}
\usepackage{array}
\usepackage{url}
\usepackage{color}
\usepackage{float}
\usepackage{wrapfig}

\title{\LARGE \bf
Neural Path Planning: Fixed Time, Near-Optimal Path Generation via Oracle Imitation 
}

\author{Mayur J. Bency, Ahmed H. Qureshi, and Michael C. Yip, \textit{IEEE Member}
\thanks{The authors are with the Department of Electrical and Computer Engineering, University of California, San Diego, La Jolla, CA 92093 USA. {\tt\small \{mbency; a1quresh; yip\}@ucsd.edu}}%
}

\begin{document}

\maketitle
\thispagestyle{empty}
\pagestyle{empty}

\begin{abstract}
Fast and efficient path generation is critical for robots operating in complex environments. This motion planning problem is often performed in a robot's actuation or configuration space, where popular pathfinding methods such as A*, RRT*, get exponentially more computationally expensive to execute as the dimensionality increases or the spaces become more cluttered and complex. On the other hand, if one were to save the entire set of paths connecting all pair of locations in the configuration space a priori, one would run out of memory very quickly. In this work, we introduce a novel way of producing fast and optimal motion plans for static environments by using a stepping neural network approach, called OracleNet. OracleNet uses Recurrent Neural Networks to determine end-to-end trajectories in an iterative manner that implicitly generates optimal motion plans with minimal loss in performance in a compact form. The algorithm is straightforward in implementation while consistently generating near-optimal paths in a single, iterative, end-to-end roll-out. In practice, OracleNet generally has fixed-time execution regardless of the configuration space complexity while outperforming popular pathfinding algorithms in complex environments and higher dimensions\footnote{Media (pictures, video) and supplemental analysis are immediately available online at \url{https://github.com/mayurj747/oraclenet-analysis} \label{footnote_1}}.
\end{abstract}

\IEEEpeerreviewmaketitle

\section{Introduction}

Being able to come up with a quick and accurate motion plan is critical to robotic systems. Motion planning involves finding a connection between two locations while avoiding obstacles and respecting bounds placed on the robot's movement. The majority of work in solving the motion planning problem involves online computation of graphical or grid search strategies that scale poorly with dimensions, or sampling based strategies that scale better with dimensions but are highly dependent on the complexity of the environments. Furthermore, a fundamental trade-off has existed with available algorithms --- the trade-off between finding the optimal solution and finding a feasible solution quickly. 

\label{sec:intro}

   \begin{figure}[t]
     \centering
     \includegraphics[width=\linewidth,clip,trim={10mm 8mm 10mm 10mm}]{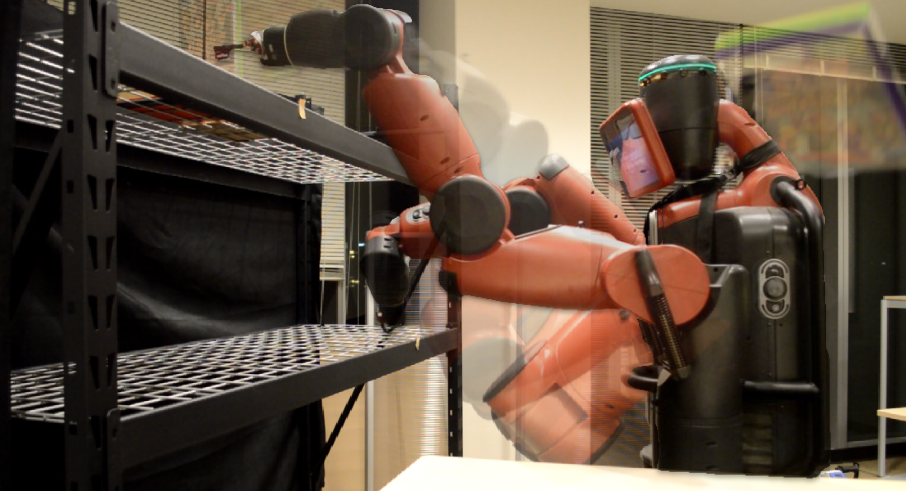}
   \caption{A composite image of the Baxter robot in action. The robot is shown executing a sequence of joint commands generated by OracleNet for picking up objects from the shelf. }
   \label{real_robot}
   \end{figure}

The main contribution of this paper is a novel approach to the general motion planning problem that leverages a neural-network based path generator, and produces feasible paths in fixed time that mimic an oracle algorithm (one that can always generate the optimal paths across the entire configuration space of the robot/environment, for any start or end goal). Our approach leverages the Recurrent Neural Network (RNN) in order to mimic the stepwise output of an oracle planner in a predefined environment, moving from the start to the end location in a relatively smooth manner. Several important advantages of OracleNet that are demonstrated in this paper: (1) \textit{it generates extremely fast and optimal paths online}; (2) \textit{it offers a valid path if one exists with probabilistic completeness}; (3) \textit{it has consistent performance regardless of the configuration space complexity}; and (4) \textit{it scales almost linearly with dimensions}. We demonstrate the results of our method on a point-mass robot, 3, 4, 6, and 7-degrees of freedom (DOF) robots. Because our algorithm scales close to linear with increase in dimensions, making it significantly faster and more efficient for online motion planning for non-trivial problems, compared to the polynomial or worse time complexity of popular motion planners. 

Training of OracleNet is required on each new environment, making it suitable for complex robots operating in a largely static space (such as a warehouse, shopping mall, airport terminal, etc.). Small, local perturbations are allowable given retrain and repair strategies that retain probabilistic optimality and completeness guarantees. Although not investigated in this work, OracleNet can also be made to adapt to dynamic environments by contextualizing path generation on obstacle representation in addition to the start and goal. This is considered in \cite{qureshi2018motion}, where we expand on the original concept of neural motion planning discussed in this paper to generate paths efficiently in a variety of unseen environments and compares against the state-of-art planners. However, encoding environment information, especially for multi-DOF systems, is far from trivial \cite{qureshi2018motion} because 1) the robot's surrounding environment and joint-space constitute entirely different distributions and therefore mapping from one to the other is challenging and a research problem of its own; 2) the environment information is usually taken as raw point-cloud data, encoding which into an interpretable feature space is hard as the representation needs to respect the permutation invariance of the input point cloud. OracleNet implicitly learns an obstacle encoding from its training set of expert trajectories, thereby by-passing expensive environment encoding. An active learning strategy may also be employed that will enable OracleNet to be retrained to incorporate changes in the environment in just a few iterations. In general, this approach offers the starting concept of formulating the path planning problem with a sequential neural network-based solver (i.e. \textit{neural motion planning}) in which many algorithmic variants can be considered, including those that extend to unseen environments and dynamic environments.

\section{Related Work}
A range of techniques to solve the motion planning problem has been proposed in the past two decades, from algorithms that emphasize optimality over computational expense, to those that trade-off computational speed with optimality \cite{rickert2008balancing, qureshi2016potential, qureshi2015intelligent}. Traditional algorithms such as A* \cite{hart1968formal} that search on a connected graph or grid, while fast and optimal on small grids, take exponentially longer to compute online with increasing grid sizes and environment complexity. Since the results presented in this work focus on static environments, precomputed roadmap-based planners can also be considered. Probabilistic Roadmaps (PRM), while appearing to solve the sample-size issue by decoupling the number of samples required to construct a graph from the dimensionality of the space, suffer from the same fate of requiring an exponential number of points in dimension to maintain a consistent quality. In fact, any sampling scheme, random or deterministic, has been shown to suffer from this \cite{lavalle2004relationship}. Preliminary results for benchmarking PRM have been provided in \ref{footnote_1}. Sampling-based strategies such as RRT have better computational efficiency for searching in high dimensional spaces \cite{lavalle1998rapidly} but get slowed down by their ``long tail" in computation time distribution in complex environments. Apart from using grid-based and sampling-based motion planners, optimizing over trajectories has also been proposed, with approaches such as using potential fields to guide a particle's trajectory away from obstacles \cite{khatib1986real} and reformulating highly non-convex optimization problems to respect hard constraints \cite{ratliff2009chomp}. A review of recent algorithms and performance capabilities of motion planners can be found in \cite{gonzalez2016review}.

The challenge of creating and optimizing motion plans that incorporate the use of neural networks has long been a problem of interest, though computational efficiency in solving for deep neural networks has only recently made this a practical avenue of research. An early attempt aimed to link neural networks to path planning by specifying obstacles into topologically ordered neural maps and using neural activity gradient to trace the shortest path, with neural activity evolving towards a state corresponding to a minimum of a Lyapunov function \cite{glasius1995neural}. More recently, a method was developed that enables the representation of high dimensional humanoid movements in the low-dimensional latent space of a time-dependent variational autoencoder framework \cite{chen2016dynamic}.

Reinforcement Learning (RL) approaches have also been proposed for motion planning applications \cite{levine2013guided, silver2017mastering}. Recently, a fully differentiable approximation of the value-iteration algorithm was introduced that is capable of predicting outcomes that involve planning-based reasoning \cite{tamar2016value}. However, their use of Convolutional Neural Networks to represent this approximation limits their motion planning to only 2D grids, while generalized motion planning algorithms can be extended to arbitrary dimensions. RL assumes that the problem has the structure of a Markov Decision Process where the agent attempts to solve the problem through a trial-and-error based interaction with the real environment. On the other hand, classical motion planning algorithms take in a full state information map as a part of the planning problem and output a solution without a single interaction with the real environment. The algorithm presented in this work leverages the assumptions used in the latter and thus is different from RL-based approaches.  

Learning to generate motion plans has also been considered via a Learning-from-Demonstration (LfD) approach. Using an “expert” (usually human) to provide demonstrations of desired trajectories, LfD methods are able to generalize within the set of demonstrations an approximate, underlying sequence or policy that reproduces the demonstrated behavior. LfD has been successfully applied in various situations that involve challenging dynamical systems or nuanced activities \cite{abbeel2010autonomous, calinon2010learning}. Our path generating algorithm can be considered an extension of LfD since the lines between motion planning, imitation learning, and model predictive control are getting blurred with advancements in machine intelligence. 

\begin{figure}[t]
      \centering
       \includegraphics[scale=0.5]{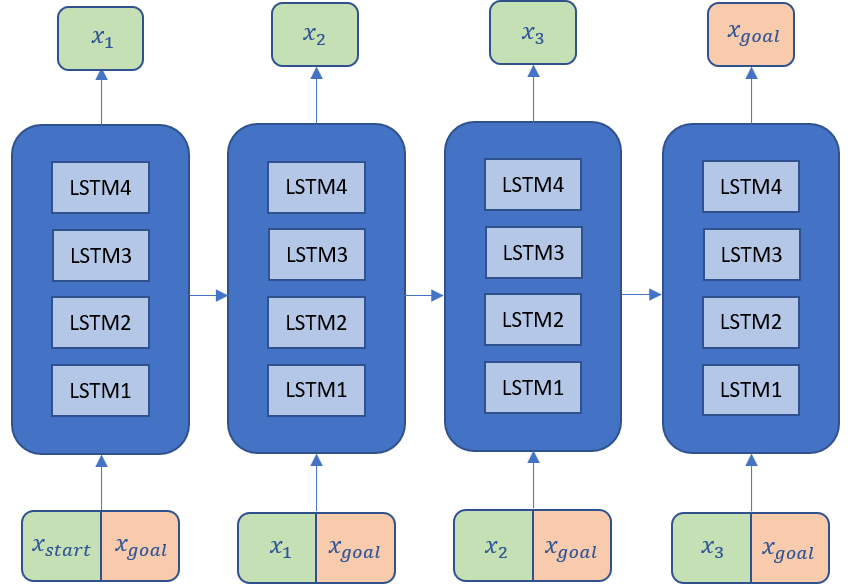}
      \caption{An ``unfolded through time" strategy for motion planning is proposed in OracleNet. Note that within this RNN structure, the Long Short Term Memory (LSTM) hidden layer weights are shared across timesteps as the inputs get iteratively updated and concatenated.}
      \label{fig1}
      \vspace*{-0.2in}
   \end{figure}

\section{Methods}
\label{sec:methods}
\subsection{Problem Definition}
In this work, we use the standard definition of configuration spaces (c-spaces) to construct the environment in which our motion planning algorithm operates. For a robot with $d$ degrees-of-freedom (DOF), the configuration space represents each DOF as a dimension in its coordinate system. Each $d$-dimensional point in the c-space represents the $d$ joint angles of the robot and therefore, the full configuration of the robot in the real world. Due to this property, motion planning in c-spaces is simpler than in geometric spaces. A motion planning task involves connecting two points in a $d$-dimensional c-space, with obstacles mapping from Cartesian space to c-space in a nonlinear fashion through forward-kinematic collision checks \cite{das2017fastron}. A necessary assumption for our algorithm is knowledge of the configuration space, which is defined by a given robot present in a given static constrained environment. For example, these problems arise in solving for optimal routes in a large network of roads or navigating through the lanes of a crowded warehouse.

Let $X \subset \mathbb{R}^d$ be the c-space. Let $X_{obs} \subset X$ be the obstacle region, such that $X \backslash X_{obs}$ is an open set, and denote the obstacle-free space as $X_{free} = cl(X \backslash X_{obs})$, where $cl(·)$ denotes the closure of a set. The initial start point $x_{start}$ and the goal $x_{goal}$, both elements of $X_{free}$, are provided as query points. The objective here is to find a collision-free path between $x_{start}$ and $x_{goal}$ in the c-space. Let $x$ be defined as a discrete sequence of waypoints. $x$ is considered valid if it is continuous, collision free (each waypoint in the generated path should lie in $X_{free}$), and feasible $(x(0) = x_{start}, x(t) = x_{goal}$ for some finite \textit{t}).  

\subsection{Proposed Algorithm}
We propose to solve the problem of generating goal-oriented path sequences by passing in a goal location as an auxiliary input at each step of the prediction, which provides a “reminder” to the network about where it should ultimately converge. At each step, the input vector is concatenated with the desired goal location and the resulting augmented input vector is used to train an RNN model. The RNN is trained on optimal trajectories that span the entire configuration space, which allows it to internalize an oracle-mimicking behavior that generates optimal path sequences during rollouts. The network comprises of stacked Long Short Term Memory (LSTM) layers that preserve information over a horizon of outputs \cite{hochreiter1997long}, with the output layer being fully connected to the final LSTM hidden layer. In this work we concern ourselves with fixed short-term memory; our experiments showed that using LSTMs resulted in better performance over vanilla RNNs.  Mean Squared Error between the predicted output and the teaching signal is used to train the network. The number of layers was decided on by empirically converging to an appropriate size based on the dimensionality of the problem to be solved. Increasing the number of layers as we increase the dimension of the c-space to capture additional degrees of freedom in the training set lead to better performance. For any finite time trajectory of a given $n$-dimensional dynamical system, a practical limit in expanding network size can be approximately realized by the internal state of the output units of a continuous time recurrent neural network with $n$ output units, some hidden units, and an appropriate initial condition \cite{funahashi1993approximation} . Exact numbers for network size and depth are provided in the Section \ref{sec:results}.

\subsection{Training Set Creation and Offline Training}
A training set consisting of a number of valid paths created by an ``expert" planner, which we call the Oracle. We used the A* to generate an optimal set of paths for training.  The c-space is sampled to create a graph with connected nodes. Two nodes are randomly selected without replacement (based on a uniform distribution) from the set of nodes present in $X_{free}$ and A* is executed to find the optimal path connecting them. This process is repeated $N$ times to obtain a training set consisting of $N$ ``expert'' paths. Each generated path is split into their composing waypoints to make each individual waypoint represent a sample in the training set. The teaching signal corresponding to each sample then becomes the next waypoint in the path sequence. That is, if $x$ is a path with $\tau$ waypoints, the path is split into \{$x(0)$, $x(1),...,x(\tau-1)$\} and the corresponding teaching signals become \{$x(1)$,  $x(2),...,x(\tau)$\}. If we assume that $x(\tau)$ is the goal point in a sequence, then the auxiliary input is concatenated as $\tilde{x}(t) = [x(t), x(\tau)]$, where $t$ ranges from 0 to $\tau$.  The LSTM network is trained on the $\tilde{x} \forall t,N$.

\subsection{Online Execution through Bi-directional Path Generation}

\begin{figure*}%
    \centering
    \includegraphics[width=0.32\linewidth,clip,trim={10mm 8mm 6mm 10mm}]{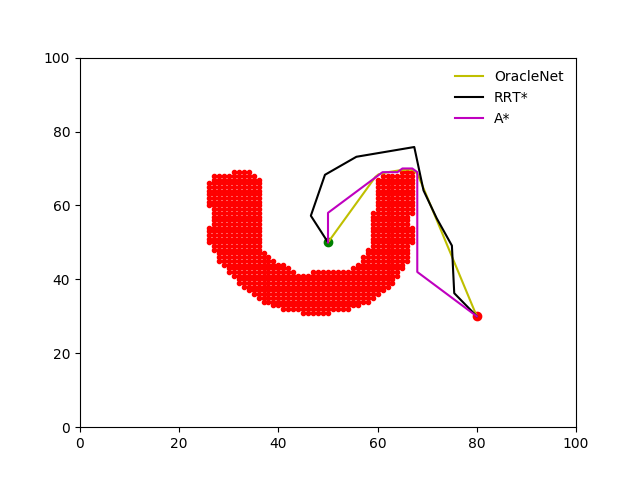}
    \label{example_fig}
    \includegraphics[width=0.32\linewidth,clip,trim={10mm 8mm 6mm 10mm}]{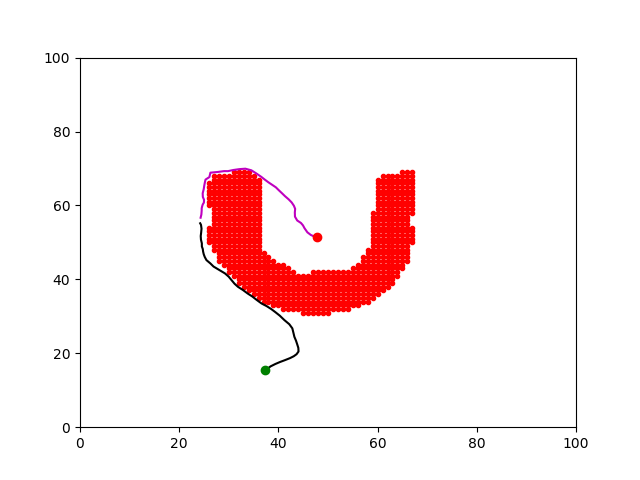}\label{bidirectional_example}
    \includegraphics[width=0.32\linewidth,clip,trim={10mm 8mm 6mm 10mm}]{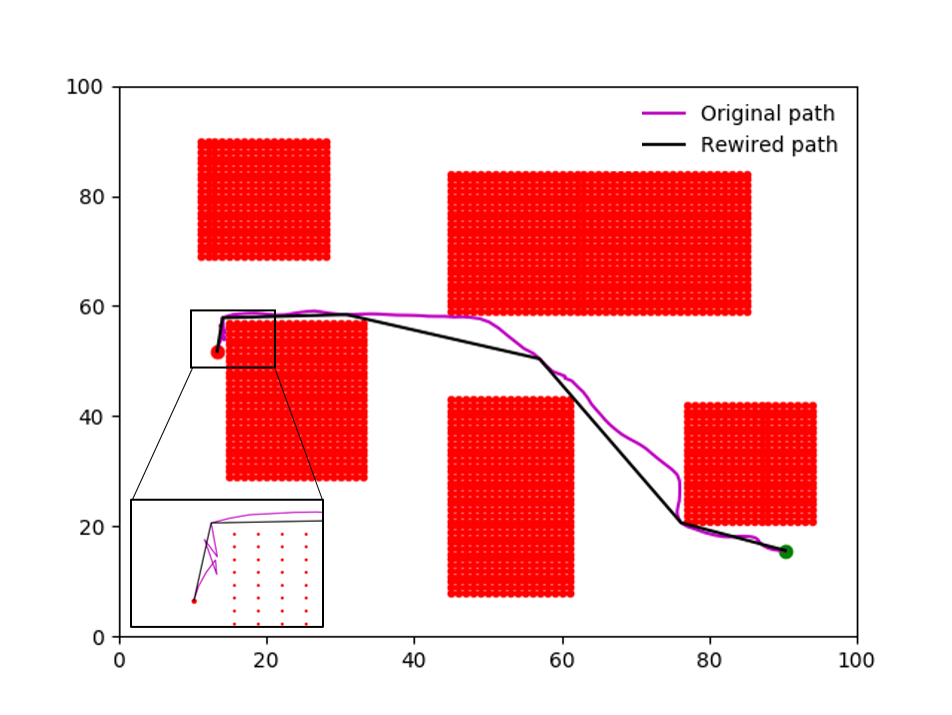}   
    \caption{(a) An example path generated by OracleNet, A*, and RRT*. (b) Example of bi-directional path generation (not rewired) between start (red) and goal (green) configurations. (c) Example of bi-directional, repaired, and rewired path generation. Figure inset shows the repair module in action (magenta line). After the path is repaired and converges successfully, the rewire module is called to remove superfluous nodes and any kinks the repairing may have introduced (black line).}\label{fig3}    
\vspace*{-0.2in}\end{figure*}

Given a trained network, for testing, we select two points $y_{start}$ and $y_{goal}$ (not from the training set) from $X_{free}$ and attempt to roll out a path connecting them while avoiding obstacles. The network generates a sequence of waypoints until a final connection to $y_{goal}$ occurs to complete the path. After the process is terminated, all the sequentially generated outputs are formatted as waypoints in the generated path. To make the path generating process more robust, bi-directional path generation is used. We start the generation process from  $y_{start}$ and $y_{goal}$ points simultaneously and make the two branches grow towards each other. The process is terminated when the two branches meet, and the branches are then stitched together to form a complete path. Fig. \ref{fig3} (b) shows the bidirectional stepping behavior. Instead of forcing the path to ``grow" towards an arbitrarily selected fixed goal point, the network now has the option to target a constantly shifting goal point (the current point of the other branch). This increases the chance of the convergence point falling with the Oracle paths on which OracleNet was trained, thus increases feasibility and success rates, while having no impact on path roll-out time. 

\subsection{Repair and Rewire}
It is to be expected that OracleNet will never yield an exact duplication of an Oracle's behavior. Also, since the expert demonstration paths are sampled randomly on the c-space, there may be regions in the c-space that the network has never seen in the dataset. Because of these unseen regions, the generated waypoints may not adhere strictly to obstacle boundaries and cut corners through obstacle regions at times. This in practice happens less than 3\% in all paths generated by the network. Naive methods may exist such as obstacle padding. However, we propose a \textit{repair} module to fix violating waypoints as they appear online while generating the path. The repairing strategy used is straightforward to implement. When current waypoint $x(t)$ is generated inside an obstacle region, a direction is randomly selected at a step distance $\epsilon$ from \textit{x(t-1)} reach a new candidate $x_{new}(t)$. Random samples are taken until the first feasible $x_{new}(t)$ is found.  $x_{new}(t)$ then replaces $x_{t}$ and then OracleNet continues. Since the random search is initiated from $x(t-1)$ which, by definition, is a valid point, the chances there will be at least one $x_{new}(t)$ from which OracleNet can proceed are very high. With an appropriately chosen $\epsilon$ and testing the connectivity of two points with a collision checker (analogous to the ``local planner'' in PRM), feasible paths are generated in probabilistic completeness.

The repair methods above, along with general network noise, will result in paths that can be non-smooth. To deal with this, we propose a \textit{rewiring} process that removes unnecessary nodes in the paths by evaluating if a straight trajectory connecting two non-consecutive nodes in the path is collision-free. Similar to the rewiring algorithm used in RRT*, a lightweight implementation of this algorithm has very little processing overhead and thus can be used without a noticeable increase in path generation times. It is interesting to note that the final path thus generated and processed has a practical time complexity of $O(n)$ (as shown through results in Section \ref{sec:results}). To maintain fairness and consistency in optimality comparisons, the proposed rewiring technique is also applied to the output of A* searches in all our experiments. The full procedure for OracleNet in online path generation is outlined in Algorithm \ref{algo:1}. 

\begin{algorithm}[t]
\DontPrintSemicolon 
$\textbf{procedure }OracleNet(x_{start},x_{goal})$\;
$G \gets TrainedNetwork()$\Comment{Trained OracleNet}\;
$x_{current} \gets x_{start}$\;
$path \gets []$\;
\While{$x_{current} \not= x_{goal}$} {
$x_{current} \gets G([x_{current}, x_{goal}]) $\;
\If{$Obstacle(x_{current})$}
   {
   $x_{current} \gets Repair()$\;
   	  
   }
	$path.append(x_{current})$\;	
   }
	$path \gets Rewire(path)$\;
   \Return{$path$}\;

\caption{OracleNet (Online Path Rollout)}\label{oraclenet}
\label{algo:1}
\end{algorithm}

\section{Results}
\label{sec:results}

\begin{figure*}[h]
      \centering
      \includegraphics[scale=0.52]{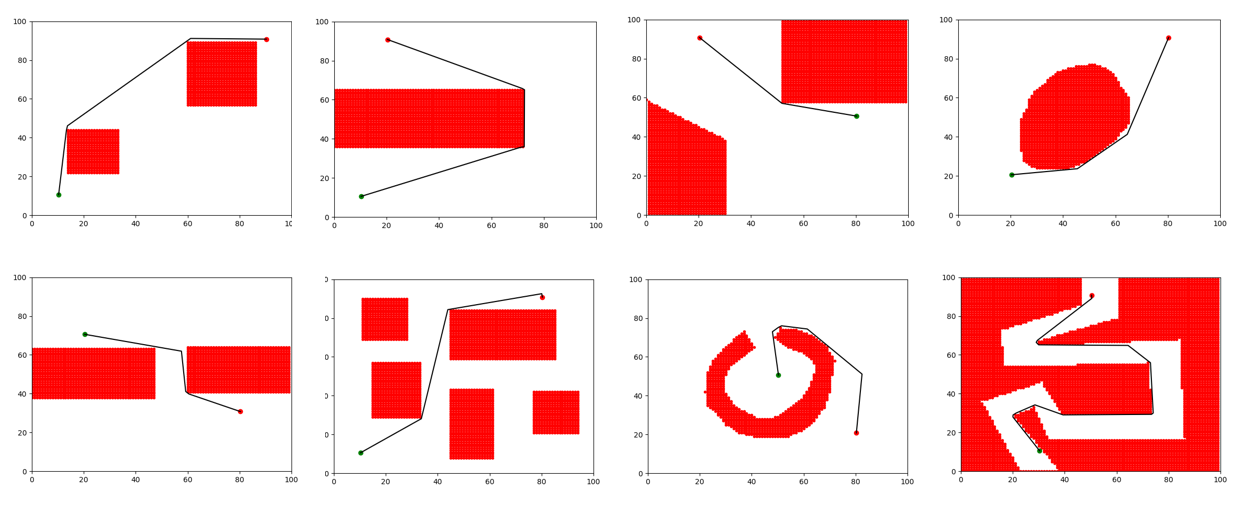}
      \caption{8 environments are used in the 2D Gridworld experiments. The top row has ``simple" environments, numbering from the left, while the bottom row has ``difficult" environments. Each environment shows a single roll-out of OracleNet. Unlike pathfinding algorithms, no expanding search is required.}
      \label{8env}
\end{figure*}

\begin{table*}[h]
\caption{Speed and Optimality of OracleNet benchmarked against A* and RRT* in a 2D environment}
\label{table_time}
\small
\begin{center}
\begin{tabular}{|c||c||c||c||c||c|}
\hline
& \multicolumn{3}{|c||}{Completion Time}  &  \multicolumn{2}{|c|}{Optimality (Ratio of Path Lengths${}^+$)}\\
\hline
Environment & A* (s) & RRT* (s) & OracleNet (s)& OracleNet / A*  & OracleNet / RRT* \\
\hline
Simple 1  & 0.08 (0.06) & 1.98 (3.68) &  0.13 (0.18)& 0.97 (0.08) &  0.84 (0.16)\\
Simple 2 &0.09 (0.07)&	1.23 (1.76)&	0.24 (0.18)&	0.96 (0.03)&	0.86 (0.11)\\
Simple 3  & 0.07 (0.06)& 0.93 (2.9) & 0.16 (0.19)& 0.96 (0.04) & 0.86 (0.10)\\
Simple 4 & 0.08 (0.05)&1.53 (2.43) &0.18 (0.20)&0.98 (0.05)&		0.87 (0.12)\\
Difficult 1&0.07 (0.05)&2.69 (3.54)&0.18 (0.10)&0.97 (0.07)&		0.87 (0.10)\\
Difficult 2& 0.07 (0.05)&3.67 (6.31)   & 0.18 (0.12)&  0.99 (0.05)& 0.88 (0.12)\\
Difficult 3&  0.09 (0.06) & 5.17 (11.57) & 0.17 (0.12) & 0.96 (0.22) & 0.87 (0.12)\\
Difficult 4& 0.05 (0.04)& 8.79 (12.81) & 0.18 (0.09)& 0.96 (0.17) & 0.89 (0.11)\\
\hline
\end{tabular}

\vspace{3pt}
\footnotesize{\textit{Values are listed as ``mean (standard deviation)'' for 8 different environments.}\\\textit{{$^+$}Lower is better. Below 1 means that OracleNet produces shorter paths.}}
\end{center}
\end{table*}

To appropriately evaluate the performance and capability of OracleNet, we test it on a number of distinct environments. The experiments were conducted in a 2D Gridworld with a point robot having translation capabilities only, 3-link, 4-link, and 6-link robot manipulators. For all experiments presented here, training is accomplished with Tensorflow and Keras, a high-level neural network Python library \cite{chollet2015keras}, with a single NVIDIA Titan Xp used for GPU acceleration.
\subsection{2D Gridworld}
Fig. \ref{8env} shows snapshots of 8 environment examples, 4 considered ``simple'' environments for popular motion planning strategies such as RRT*, and 4 ``difficult'' environments. An environment is considered simple if the obstacles are convex and widely spaced apart, while difficult environments consist of either a large number of obstacles or highly non-convex obstacles forming narrow passageways. Each continuous space environment is 100 units in length and width. A*  is run on a unit grid of 100 x 100 on this environment to generate 20,000 valid Oracle paths for training OracleNet. This set was split in accordance to the 80-20 rule, with 20$\%$ being kept for testing to control overfitting. The network architecture consists of 4 LSTM layers each of 256 hidden units. 

\begin{figure*}[h!]
      \centering
      \includegraphics[width=1.0\linewidth,clip,trim={2mm 3mm 0 0}]{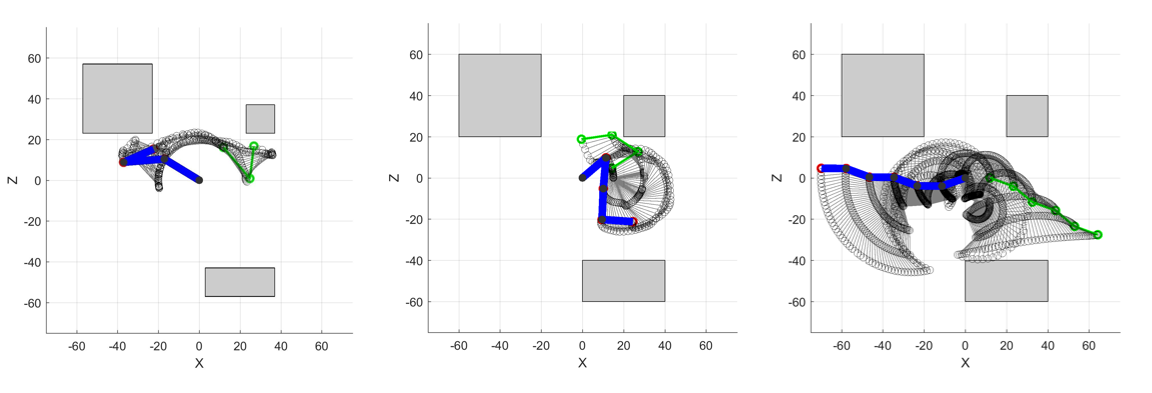}
      \caption{Example plots of successful paths generated on 3-link, 4-link, and 6-link robots (from the left) using our algorithm, showing the trail of the links as it makes its way from its initial joint configuration (marked as green). }
      \label{fig5}
   \end{figure*}

To benchmark our algorithm’s performance with existing motion planning algorithms, we use RRT* and A*. While these planners are not among the most efficient (and we do compare neural planning against the state of art in \cite{qureshi2018motion}), they are both widely used and accessible, and demonstrate the advantages of training a neural planner over using the planning algorithms during runtime. Three performance metrics are used: success rate, roll-out time, and path optimality. A generated path is considered successful if none of the waypoints encroach into obstacle region. Roll-out time measures the time taken for the network to generate waypoints from start to goal (or, in the bidirectional case, the time taken for both branches to meet). Path optimality is simply the fraction of the path length generated by OracleNet when benchmarked against paths generated by A* and RRT* respectively.  1000 randomly initialized trials were conducted for each of the 8 environments. Table \ref{table_time} shows that OracleNet manages to be comparable to A* and faster than RRT* even for a small 2D grid while being slightly more optimal in both cases. This is due to the rewiring module and the network being able to generate points in continuous space as opposed to being restricted to a discrete grid in A*. Thus, the proposed algorithm may be a suitable alternative even for small grids with limited connectivity, if path optimality is the priority. 

\begin{figure}[h]
\centering
\includegraphics[width=0.95\linewidth,clip,trim={6mm 3mm 14mm 10mm}]{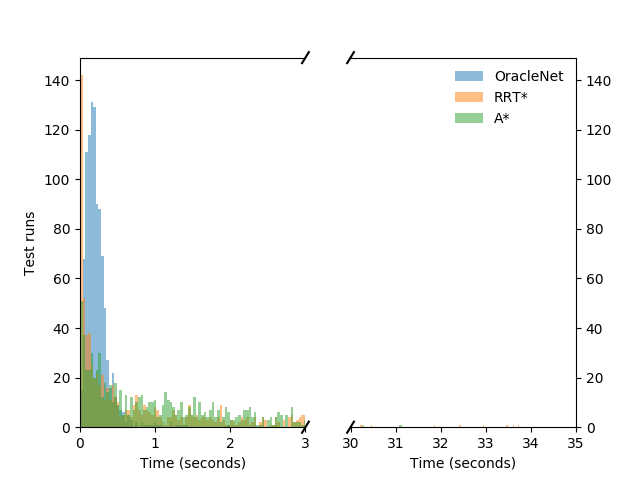}
\caption{A histogram of the test cases used to evaluate performance on a 3-link arm. Refer Table \ref{table_time_3d} for means and standard deviations. Note the Gaussian shape of the distribution for OracleNet, compared to the left-skewed exponential distributions of A* and RRT*. Higher standard deviations for A* and RRT* cause the generation times to have a much wider spread, while OracleNet's much tighter spread indicates its consistency in near fixed-time execution.}
\label{fig6}
\end{figure}

\begin{table*}[t]
\small
\caption{Performance of RNN-based motion planners in higher dimensions benchmarked against A* and RRT*}
\label{table_time_3d}
\begin{center}
\begin{tabular}{|c||c||c||c||c||c|}
\hline
& \multicolumn{3}{|c||}{Completion Time}  &  \multicolumn{2}{|c|}{Optimality (Ratio of Path Lengths${}^+$)}\\
\hline
Environment & A* (s) & RRT* (s) & OracleNet (s)& OracleNet / A*  & OracleNet / RRT* \\
\hline
3-link (3D)  & 2.707 (4.28)  & 3.83 (6.68)   &  0.22 (0.21)    & 1.02 (0.12)    & 0.75 (0.20)     \\
4-link (4D)  & 61.87 (95.07) & 18.21 (14.76) &  1.18 (0.87)    &	0.99 (0.14)    & 0.86 (0.11)     \\
6-link (6D)  & 727.56 (105.11)            & 29.32 (6.25)  &  1.24 (0.72)    &  0.95 (0.15)              & 0.85 (0.17)    \\
\hline
\end{tabular}

\vspace{3pt}
\footnotesize{\textit{Values are listed as ``mean (standard deviation)''.}\\
${}^+$Below 1 means that OracleNet produces shorter paths.}
\end{center}
\end{table*}

\subsection{Multi-Link Planar Manipulators}

To demonstrate the extensibility of the proposed method to higher dimensions, OracleNet was tested on 3-link, 4-link, and 6-link manipulators. The base link has movement range 0 to 2$\pi$ while the subsequent links can move between $-\pi$ to $\pi$. For the 3-link manipulator, we discretized the 3-dimensional joint angle c-space into a 3D uniform grid with 50 nodes on each axis, resulting in a total of $50^3$ = 125,000 uniformly spaced nodes. For the 4-link and 6-link cases, the corresponding $n-$dimensional c-space is discretized into 40 and 10 uniformly spaced nodes per axis, respectively. As in the 2D case, A* is used to generate the training set. To get an accurate representation of the complete c-space in the training set (which directly correlates to the increased number of nodes in the grid used here for training), we use 400,000 for the 3-link case, and 1 million each for the 4-link and 6-link cases. Keeping in mind the increased number of dimensions to learn, we updated the architecture to have 6 layers with 256 units each for the 3-link case, and  6 layers with 400 units each for both 4-link and 6-link cases. For our performance evaluation, we randomly generate 1000 pairs of start and goal locations in continuous c-space. Paired with the repair and rewire modules discussed in the previous sections, we observed a 100$\%$ success rate in finding feasible paths. Generation times and path optimality is benchmarked against A* and RRT*. Table \ref{table_time_3d} shows that OracleNet scales much better than A* and RRT*. An interesting observation to note is the low standard deviation of path generation times for OracleNet. This is further expanded on in the histogram of the test cases shown in Fig. \ref{fig6}. This is indicative of the consistency of OracleNet in producing its paths across the entire c-space, whereas A* and RRT* are heavily influenced by the relative locations of the query points and the obstacles. More about this is discussed in Section \ref{sec:discussion}. 

\subsection{7-DOF Dual-Arm Baxter Robot}
\begin{figure}
    \centering
    \includegraphics[scale=0.2]{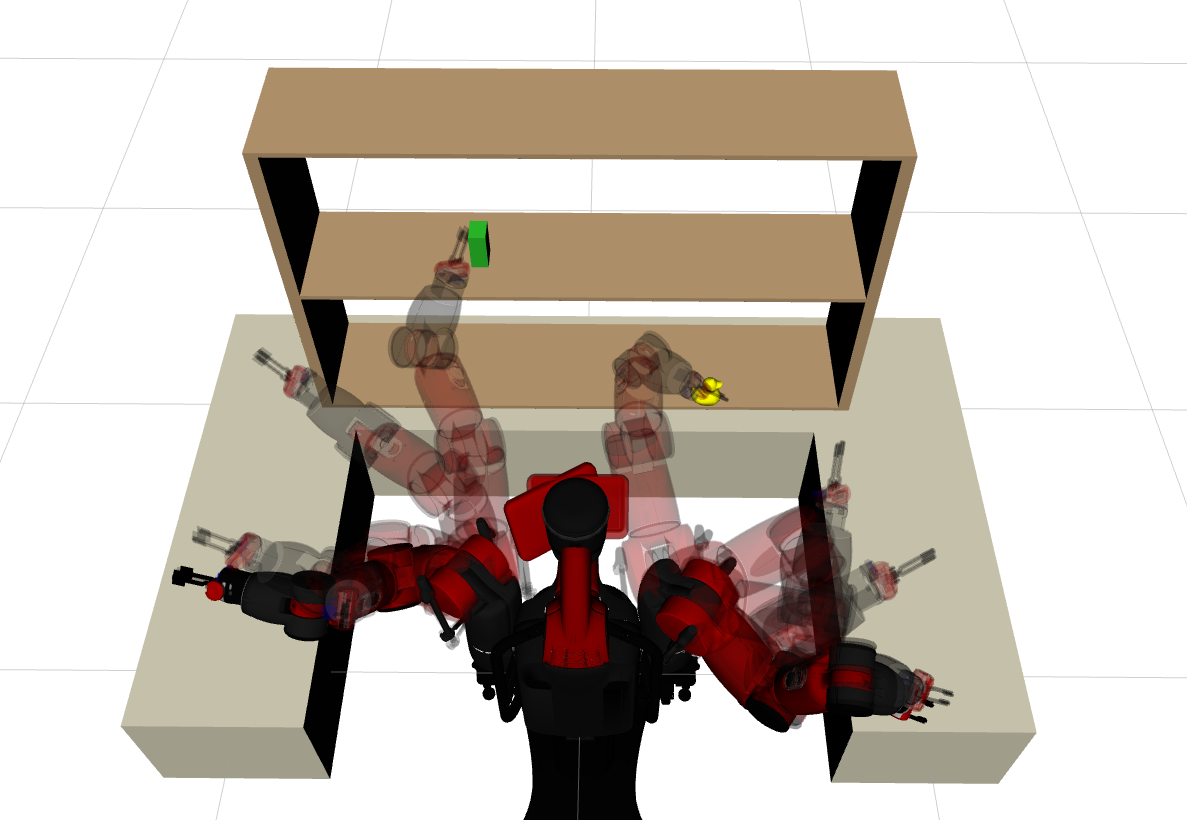}%
    \caption{A composite image of the Baxter executing a pick-and-place task with manually controlled grasp. The trail of the arms show the path taken by them to reach their respective objects. The right arms plans towards the duck while the left arm plans towards the green cuboid.}%
    \label{baxter_task_1}%
\end{figure}

As a final demonstration of OracleNet's capabilities, we show its application on a humanoid dual-arm Baxter robot.
We train the network to output a sequence of 7 DoF joint-angles (for one arm) for the robot to follow from a starting configuration to a goal configuration. The robot, after training, is expected to generate a sequence of feasible configurations (if they exist) connecting given start and goal configurations. Due to the high dimensionality of the configuration space, it becomes infeasible to discretize it to a grid of reasonable resolution and use graph search algorithms such as A* to generate optimal paths. Instead, we rely on Open Motion Planning Library (OMPL) to generate a set of paths between randomized valid configurations using RRT-Connect \cite{kuffner2000rrt}. RRT-Connect is an efficient sampling-based planner that combines RRT's sampling scheme with a simple greedy heuristic to generate quick single-query paths. Fig. \ref{baxter_task_1} shows a simulated version of the robot performing a pick-place-task. OracleNet generated the sequence of joint angles necessary for the robot arms to move towards the given object locations. Fig. \ref{real_robot} shows this being achieved with the physical Baxter robot as well. A relatively small dataset of 40,000 paths is generated and train on it using a model with a reduced number of layers (2 and 3 layers with 400 units each were experimented with). Due to the significantly constrained work-space in which the Baxter has to operate, connecting arbitrary configurations for testing may not always lead to viable paths. Even RRT-Connect, the ``expert planner" used here, frequently failed to generate viable paths before timing-out. Despite the lower-quality dataset available for this experiment, utilizing its generalizing ability and the fail-safe measures described previously, OracleNet was able to generate smooth paths connecting the chosen configurations with mean and standard deviation of generation times as 0.56 and 0.39 seconds respectively.
 
\subsection{Scalability and Effects of Path Length}
Fig. \ref{fig8} shows the scalability of OracleNet with dimensions. In conjunction with Table \ref{table_time}, it shows that OracleNet provides minimal improvements to readily accessible algorithms like A*  in low dimensional cases with small grid sizes (low resolution). In spaces with dimensions greater than 2, OracleNet scales far better, as A* (paired with a Euclidean distance heuristic) scales exponentially worse with dimensions, and RRT* (depending on the trade-off selected between step size and path optimality, scales polynomial at best). Results presented in Table \ref{table_time_3d} and Fig. \ref{fig8} present the trends that show a modest increase in execution time for OracleNet while path optimality rivals A* and far outstrips RRT*. 
   
\begin{figure}[]
      \centering
      \includegraphics[width=0.95\linewidth,clip,trim={7mm 8mm 10mm 10mm}]{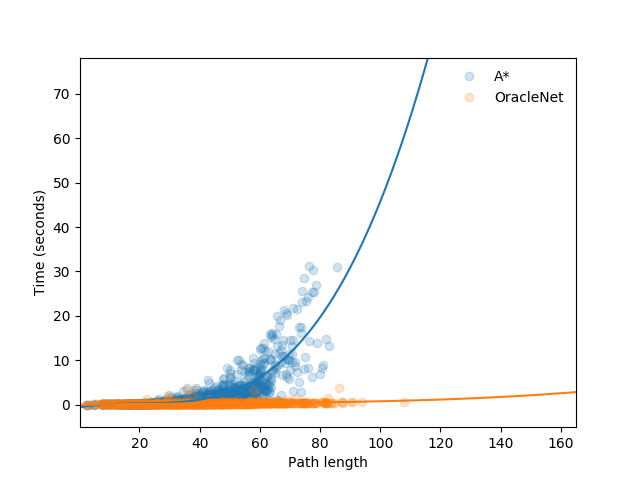}
      \caption{OracleNet and A* path generation times plotted as a function of path lengths for the 3-link robot presented here. Note the near fixed-time generation times for OracleNet, irrespective of where the query points are located in the entire c-space.}
      \label{fig9}
   \end{figure}

\begin{figure}[]
      \centering
      \includegraphics[width=0.98\linewidth,clip,trim={10mm 6mm 10mm 10mm}]{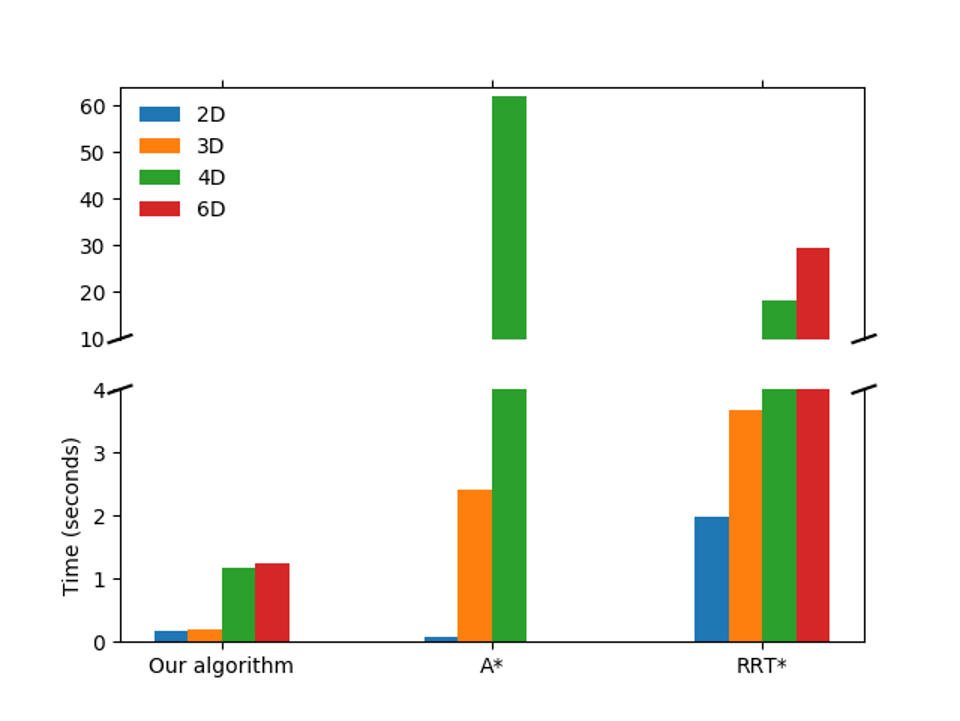}
      \caption{A comparison of average path generation times for our algorithm, A*, and RRT* for all the cases presented here. Because of the unreasonably high complexity of running A* in a 6-dimensional c-space with full connectivity, it was chosen not to be included. The scalability for online path generation is far better for OracleNet and produces near-optimal paths through the entire space which cannot be done practically with other competing methods.}
      \label{fig8}
   \vspace*{-0.2in}
   \end{figure}

Fig. \ref{fig9} shows the spread of generation times of paths when arranged as a function of path length (demonstrated on the 3-link arm). OracleNet has a linear relationship to how far away the query points are located, meaning that stepping through the environment is done consistently with a fixed time. For pathfinding algorithms such as A*, the farther apart the query points are, the more chances are a number of nodes required to be explored increases exponentially (polynomial with an appropriately chosen admissible heuristic). The fixed-time behavior of OracleNet is especially valuable in time-critical scenarios where a feasible path must be known in fixed time.

\section{Discussion}
\label{sec:discussion}

The three main qualities required of any successful motion planner are feasibility, speed, and optimality. While feasibility is usually non-negotiable, speed and optimality can often be traded for each other depending on what the priorities are. This work attempts to achieve both high speed as well as optimality while simultaneously preserving feasibility. The experiments performed in the various environments presented cover a range of cases that support this effort.
  
Using a discrete uniform grid to build the training set allows us to estimate the efficiency of the dataset and the generalizing ability of the network. The fact that it is possible to successfully generate paths between two continuous free floating points when it is trained only on a sparse discrete grid indicates generalizing ability of the network. The 6-link experiment is a good example of this. Even with just 10 uniformly spaced discrete samples per axis, the network managed to find optimal collision-free paths connecting continuous points sampled arbitrarily in the c-space. 

Another effect of network size beyond the required training data directly correlate with path generation times, an increasing network size will increase the path generation time at a comparable rate. While overfitting is avoided (as described in Section \ref{sec:methods}), larger networks produce minimal performance improvements, and it is worth empirically honing it to an appropriate network size. A complementary strategy to potentially reduce network size while retaining performance and potentially improving robustness is using dropout \cite{srivastava2014dropout}.

A unique property of OracleNet is with regards to real-time execution. Unlike any other motion planning strategy (with the exception of the potential field strategies), a movement can be initiated immediately before a path through the environment is found. Because OracleNet encodes optimal behaviors, one can execute OracleNet under the belief that any step it takes will move towards the right direction. Issues such as repair and rewiring can be resolved by considering an N-step horizon, which is a typical strategy in online motion planning and specifically re-planning. Thus, in practice, one can expect that a robot can move and react instantaneously once given its goal state. 

The speed and invariance of performance to environments do come at a cost, namely the creation of a dataset, and time and computation power to accurately generate paths. Taking advantage of multi-processor parallelization, for the training set sizes posted in Section \ref{sec:results}, the dataset creation times ranged from around 2 hours for the 2D grids to around 12 hours for the 6D cases. For the training times, with GPU acceleration and for the network sizes mentioned in Section \ref{sec:results}, the training times ranged from around 4 hours till convergence for the smaller 2D cases to 10 hours for the larger 6D cases. This cost is easily accepted if the algorithm is viewed as an extremely fast way to create online optimal and feasible motion plans for any start and goal state in a static environment, given that the process of creating training data and training of the network occurs offline. As mentioned previously, potential real-world applications of this algorithm include warehouse scenarios, robot picking and shelving, and custodial robots.
   
Given a foundational framework for mimicking Oracles using neural networks, several extensions can be pursued. Adapting to dynamic environments, unseen environments via transfer learning, improving sampling strategies for Oracles (i.e. training data selection) are considered in \cite{qureshi2018motion}, and can be extended towards a fast and efficient local planner component in task-based planners \cite{krontiris2015dealing}.   

\bibliographystyle{IEEEtran}
\bibliography{references}
\end{document}